\documentclass[lettersize,journal]{IEEEtran}
\usepackage{amsmath,amsfonts}
\usepackage{algorithmic}
\usepackage{algorithm}
\usepackage{array}
\usepackage[caption=false,font=normalsize,labelfont=sf,textfont=sf]{subfig}
\usepackage{textcomp}
\usepackage{stfloats}
\usepackage{url}
\usepackage{verbatim}
\usepackage{graphicx}
\usepackage{booktabs}
\usepackage{cite}
\usepackage{bm}

\usepackage{orcidlink}
\usepackage{float}
\usepackage{tikz}
\usetikzlibrary{shapes, arrows.meta, positioning, fit, backgrounds, calc}

\usepackage{clipboard} 
\usepackage{subfiles} 
\usepackage{soul} 

\newcounter{reviewer}
\newcounter{point}[reviewer]
\renewcommand{\thepoint}{P\,\thereviewer.\arabic{point}} 

\newenvironment{reply}
   {\color{blue} \medskip \textbf{Response}:\  }
   {\color{black} \medskip }

\newcommand{\shortreply}[2][]{\color{blue} \medskip \noindent \begin{sf}\textbf{Reply}:\  #2
	\ifthenelse{\equal{#1}{}}{}{ \hfill \footnotesize (#1)}%
	\color{black} \medskip \end{sf}}

 \newenvironment{changes}
   {\color{blue} \medskip \textbf{Authors' Action}:\  }
   {\color{black} \medskip }

\newclipboard{output-IEEE_Journal}

\begin{document}

\title{\textbf{ReLANCE}: A Resource-Efficient Low-Latency Cortical Neural Acceleration Engine}

\author{Sonu Kumar, Arjun S. Nair\textsuperscript{\textdagger}, Bhawna Chaudhary\textsuperscript{\textdagger},\\ 
Mukul Lokhande, \IEEEmembership{Member, IEEE}, and Santosh Kumar Vishvakarma, \IEEEmembership{Senior Member, IEEE}
\thanks{Manuscript received Month XX, 202x; revised Month XX, 202x.}

\thanks{\textsuperscript{\textdagger}Arjun S. Nair and Bhawna Chaudhary contributed equally to this work and are with the NSDCS Research Group, Dept. of Electrical Engineering, IIT Indore, India. Sonu Kumar and Santosh K. Vishvakarma thank the Dept of Science and Technology (DST), Govt of India, for the INSPIRE PhD fellowship, and MeitY/SMDP-C2S for ASIC design tools and are associated with the Centre for Advanced Electronics, Indian Institute of Technology (IIT), Indore, 453552, India. Mukul Lokhande was with the NSDCS Research Group, IIT Indore, India, and is currently with Qualcomm Technologies Inc., Bengaluru 560066, India.}
\thanks{\textbf{Corr. author}: Santosh K. Vishvakarma (e-mail: skvishvakarma@iiti.ac.in).}
}



\markboth{IEEE Transactions on Very Large Scale Integration (VLSI) Systems,~Vol.~XX, No.~X, Month~202x}%
{Kumar \MakeLowercase{\textit{et al.}}: A Resource-Efficient Low-Latency Cortical Acceleration Engine}


\maketitle
\begin{abstract}
We present a Cortical Neural Pool (CNP) architecture featuring a high-speed, resource-efficient CORDIC-based Hodgkin–Huxley (RCHH) neuron model. Unlike shared CORDIC-based DNN approaches, the proposed neuron leverages modular and performance-optimised CORDIC stages with a latency-area trade-off. \textcolor{black}{\Copy{R1.1A}{We introduce a novel Constraint-Aware Modular Parallelism (CAMP) with Precision \& Stability handling to leverage maximum speedup and utilisation of hardware through hardware software co-design.}} The FPGA implementation of the RCHH neuron shows 24.5\% LUT reduction and 35.2\% improved speed, compared to SoTA designs, with 70\% better normalised root mean square error (NRMSE). Furthermore, the CNP exhibits 2.85× higher throughput (12.69 GOPS) than a functionally equivalent CORDIC-based DNN engine, with only a 0.35\% accuracy drop relative to the DNN counterpart on the MNIST dataset. The overall results indicate that the design shows biologically accurate, low-resource spiking neural network implementations for resource-constrained edge AI applications. The reproducibility codes are publicly available at \url{https://github.com/mukullokhande99/CNP_RCHH}, facilitating rapid integration and further development by researchers.

\end{abstract}


\section{Introduction}
\begin{figure}[!t]
    \centering
    \includegraphics[width=0.875\columnwidth, height=52.5mm]{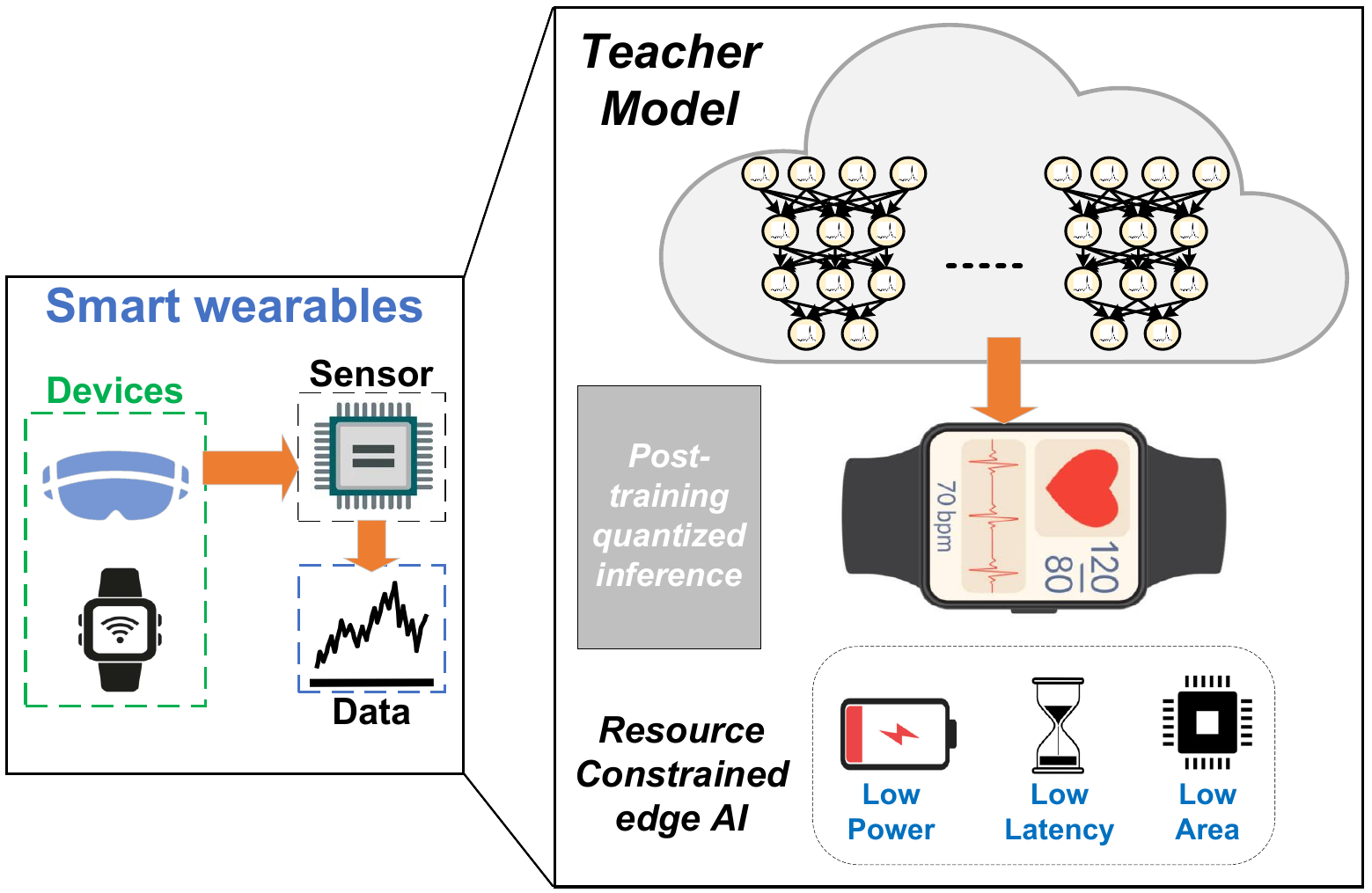}
    \caption{Visualisation depicting SNN use cases in wearable devices, emphasis on resource-constrained implementation, adapted from \cite{Neuromorphic_accl}}
    \label{fig:helicopter}
\end{figure}

\IEEEPARstart{A}{rtifical} Intelligence (AI) is a significant decision-maker in human life, especially in the era of data-driven sensory computation. Conventionally, Deep neural networks (DNNs) have eased daily human tasks; however, the rising cost has been a limiting factor due to their computational complexity and hardware resource requirements \cite{RPE, CAL'17, ESL'24}. Furthermore, the significant data dependency leads to high operational power consumption and slow event-triggered responses, making practical deployment difficult \cite{DNN_TRETS'23, Flex-PE}. Spiking Neural Networks (SNNs) have shown lower resource consumption than DNNs, with spike-based signal processing similar to that of biological neurons \cite{Neuromorphic_accl, Nature4}. The challenges associated with a hardware-software co-design approach for resource-constrained SNN implementation in smart wearables are depicted in Fig. \ref{fig:helicopter}. The hardware complexity can be gauged from ViT-G\cite{TCAD'23}, an object detection model that performs a single inference on ImageNet (224x224x3) with 2.86 GMACs and 184B parameters, requiring 3 $\times$10\textsuperscript{4} days for training and 159 MWh of energy consumption on TPUv3. This emphasises focus on 
low-power, low-latency, and low-area edge AI implementation, close to sensor/data in devices and quantised inference. 

At present, several works \cite{HH-TCASI'21, IF_TCASI'16, Izhikevich_NEWCAS'21, PWL_HH_Frontiers'14, CORDIC_Izhikevich_TBioCAS'22} focus on specialised hardware accelerators using application-specific integrated circuits (ASICs) and field-programmable gate arrays (FPGAs) with limited resources. Prior works proposed different neuromorphic hardware acceleration approaches with neurons such as Leaky Integrate-Fire (LIF)\cite{TCAD'23}, Hodgkin-Huxley (H\&H)\cite{HH-TCASI'21, PWL_HH_Frontiers'14}, and Izhikevich\cite{Izhikevich_NEWCAS'21} for spiking implementations, and for enhanced hardware efficiency, they addressed the design approach with the CORDIC approach with base-2 function\cite{IF_TCASI'16, CORDIC_Izhikevich_TBioCAS'22} or multiplier-less implementation, piece-wise linear (PWL) approximation and read-only-memory (ROM), etc. This extended to inference with low power and high parallel throughput, yielding comparable detection accuracy. Thus, the primary goal of this work is to evaluate the SNN topology, compare it with its DNN counterpart in terms of FPGA utilization, and assess application performance in a real-time environment.

\newcommand{\RoneOneB}{%
The main contributions of this brief are:
\begin{itemize}

\item \textbf{RCHH Neuron (Resource-efficient CORDIC-enhanced HH Core):}
We propose a resource-efficient CORDIC-based Hodgkin--Huxley (RCHH) neuron that replaces a shared/generic CORDIC datapath with optimal CORDIC stages and constraint-aware scheduling for rate computation. On a Xilinx VC707 FPGA, the proposed neuron achieves \textbf{24.5\%} LUT reduction and \textbf{35.2\%} speed improvement over the state-of-the-art iterative CORDIC HH design~\cite{HH-TCASI'21}, and reduces Normalised Root Mean Square Error (NRMSE) by \textbf{70\%}.

\item \textbf{RCHH-Neuron-Based Cortical Neural Pool (CNP) Engine:}
We extend the proposed RCHH neuron into a Cortical Neural Pool (CNP) engine with integrated Precision \& Stability handling and constraint-aware modular scheduling. Specifically, the controller maximises hardware utilisation by overlapping high-latency (division-intensive) and low-latency (exponential-only) rate computations, and by masking the tail latency of division stages via concurrent state-update execution. As a result, the proposed CNP achieves \textbf{2.85$\times$} higher throughput (\textbf{12.69 GOPS}) compared to a functionally equivalent CORDIC-based DNN engine~\cite{DNN_Retro'25}, while enabling low-latency neuron stepping without stalling.

\item \textbf{System-Level Empirical Evaluation and DNN Baseline Comparison:}
We implement a 64-neuron Cortical Neural Pool (CNP) engine and benchmark it against a functionally equivalent CORDIC-based DNN engine~\cite{DNN_Retro'25} on the same FPGA platform. The proposed CNP achieves \textbf{2.85$\times$} higher throughput (\textbf{12.69 GOPS}) with only a \textbf{0.35\%} drop in accuracy relative to the DNN baseline on MNIST, demonstrating the practical benefits of the SNN approach. In addition, we provide a scalability analysis up to \textbf{1024 neurons} and discuss implications for higher-resolution inputs and larger SNN model configurations. The codes are open-sourced at \url{https://github.com/mukullokhande99/CNP_RCHH}.

\end{itemize}%
}

\textcolor{black}{\Copy{R1.1B}{\RoneOneB}}

\section{Proposed ReLANCE Architecture}
\label{sec:proposed}

RELANCE refers to a deliberate revisiting of classical CORDIC architectures using a spiking-inspired processing paradigm to achieve efficient, low-latency AI acceleration for modern AI workloads.The RCHH neuron is optimally implemented as a hardware-efficient discrete version of the Hodgkin-Huxley (HH) model, proposed in 1952 by Hodgkin and Huxley to describe the mathematical behaviour of biological neurons, as follows:

\begin{equation}
\begin{cases}
I = C_M \frac{dV}{dt} + \overline{g}_K n^4 (V - V_K) + \overline{g}_L (V - V_L) \\
    + \overline{g}_{Na} m^3 h (V - V_{Na}) \\
    \frac{dm}{dt} = \alpha_m (V) (1 - m) - \beta_m (V) m \\
\frac{dh}{dt} = \alpha_h (V) (1 - h) - \beta_h (V) h \\
\frac{dn}{dt} = \alpha_n (V) (1 - n) - \beta_n (V) n 
\end{cases}
\tag{1}
\label{1}
\end{equation}

Membrane potential ($V$) is the output, and injected current ($I$) is the input, where $C_{M}$ denotes the membrane capacitance, and $g$ denotes the conductance associated with each ion channel. The variables $n$, $m$, and $h$ correspond to the gating variables controlling ion flow across the membrane. $V_{Na}$, $V_{K}$, and $V_{l}$ indicate the ion channel reversal potentials. The rate functions \( \alpha_x(V) \) and \( \beta_x(V) \) describe the transition rates of these particles between different membrane states, $x \in \{m, h, n\}$. The supporting rate equations are found in Table \ref{tab:rate_constants}.

\begin{table}[!t]
    \centering
    \caption{Rate Constants for Gating Variables}
    \label{tab:rate_constants}
    \begin{tabular}{lcc}
        \toprule
        \textbf{Gating Variable} & $\bm{\alpha_x(V)}$ & $\bm{\beta_x(V)}$ \\
        \midrule
        $m$ & 
        $\dfrac{0.1(V+40)}{1 - e^{-(V+40)/10}}$ & 
        $4.0e^{-(V+65)/18}$ \\
        \addlinespace[0.5em]
        $h$ & 
        $0.07e^{-(V+65)/20}$ & 
        $\dfrac{1}{1 + e^{-(V+35)/10}}$ \\
        \addlinespace[0.5em]
        $n$ & 
        $\dfrac{0.01(V+55)}{1 - e^{-(V+55)/10}}$ & 
        $0.125e^{-(V+65)/80}$ \\
        \bottomrule
    \end{tabular}
\end{table}

\Copy{R1.2.1}{\textcolor{black}{A rigorous evaluation was required to validate the CORDIC algorithm's effectiveness in minimizing the HH neuron's hardware footprint. We conducted simulations across varying input currents and system parameters to benchmark the CORDIC-based implementation against the standard neuron model. Table \ref{tab:param_sets} outlines these validation parameters. These distinct sets were selected to verify the CORDIC architecture's fidelity across diverse neuronal operating conditions as illustrated in (Fig. \ref{fig:spike-pattern}).}

\begin{table}[t]
\caption{HH Neuron Parameter Sets Used for Simulation and Testing}
\label{tab:param_sets}
\centering
\begin{tabular}{|l|l|l|l|}
\hline
Parameter & Set 1 & Set 2 & unit \\
\hline
$C_m$ & 1 & 1 & $\mu F/cm^2$ \\
$V_{Na}$ & 57.86 & 55 & $mV$ \\
$V_K$ & -75.76 & -110 & $mV$ \\
$V_l$ & -53.86 & -95 & $mV$ \\
$g_{Na}$ & 130 & 70 & $mS$ \\
$g_K$ & 37 & 8 & $mS$ \\
$g_l$ & 0.6 & 0.23 & $mS$ \\
\hline
\end{tabular}
\end{table}

\textcolor{black}{Thus, selecting the parameters in Table \ref{tab:param_sets} and also applying the modular cordic-based approximations for multiplication operations in the Eq.~\eqref{1} and division and exponential operations in the rate equations in Table \ref{tab:rate_constants}, we implemented the RCHH neuron to schedule and pipeline the computations optimally. The RCHH neuron orchestrates its internal data flow as seen in the FSM in (Fig. \ref{fig:RCHH-FSM}(b)). The neuron state evolves from \textit{IDLE} when \textit{start\_sim} becomes high, subsequently it goes to \textit{RATE\_CALC}, \textit{POWER\_CALC}, \textit{CURRENT\_CALC}, \textit{UPDATE\_STATE} and finally \textit{DONE}.}}

\Copy{R1.2.2}{The major focus for accelerated computing in NCP, using the RCHH neurons, has been enhancing the operations that can be pipelined and executed in parallel, while reducing redundant computation. \textcolor{black}{Unlike prior SNN implementations that utilize uniform, shared CORDIC blocks to conserve area\cite{HH-TCASI'21}, ReLANCE introduces CAMP architecture for the RCHH neuron. Through simulations, it was fixed 8 iterations for CORDIC exponential terms, while 10 for multiplication and 11 for division. This maintains Precision \& Stability, handling boundary cases.}}

\subsection{\textcolor{black}{Constraint-Aware Modular Parallelism (CAMP)}}
\Copy{R1.2.3}{Conventional shared architectures time-multiplex a single generic CORDIC unit for division, multiplication, and exponentials. While this minimizes core count, it incurs significant overhead from wide multiplexers (MUX) required to route varying inputs. ReLANCE eliminates this by deploying specialized CORDIC units, hardwired specifically for either linear or hyperbolic modes. Stripping the generic mode-switching logic reduces LUT utilization by 24.5\% despite the increased instance count.

We split the rate functions into two main categories, Firstly \textit{High-Latency Groups} (Division-Intensive): Functions like $\alpha_m$, $\beta_h$, and $\alpha_n$ require both exponential and division operations (the "a/b" format). These are computationally expensive and constitute the "Critical Path." Secondly \textit{Low-Latency Groups} (Exponential-Only): Functions like $\beta_m$, $\alpha_h$, and $\beta_n$ rely primarily on exponentials. These complete in fewer clock cycles.

We deploy six specialized CORDIC cores (exponential, multiplication and division) each to calculate all rate functions in parallel. However, rather than idling the cores assigned to the Low-Latency Group once they finish, the CPN (Central Pipeline Node) Controller dynamically re-tasks them:

\begin{enumerate}
  \item \textbf{Stage A (Concurrent Launch)}: All 6 CORDIC lanes initiate calculation of their respective $\alpha$ and $\beta$ variables simultaneously.
  
  \item \textbf{Stage B (Staggered Completion \& Reuse)}: As the Low-Latency cores complete their exponential calculations (for $\beta_m$), they effectively ``release'' their hardware resources early. Instead of waiting for the High-Latency (Division) cores to finish, the controller immediately pipelines the available results into the next computation stage: the differential gating equations ($\frac{dm}{dt}$, $\frac{dn}{dt}$, $\frac{dh}{dt}$).
  
  \item \textbf{Stage C (Latency Masking)}: The computation of the state updates ($\frac{dm}{dt}$, etc.) occurs concurrently with the final pipeline stages of the slower High-Latency rate functions.
\end{enumerate}

\begin{figure}
    \centering
    \includegraphics[width=0.875\columnwidth]{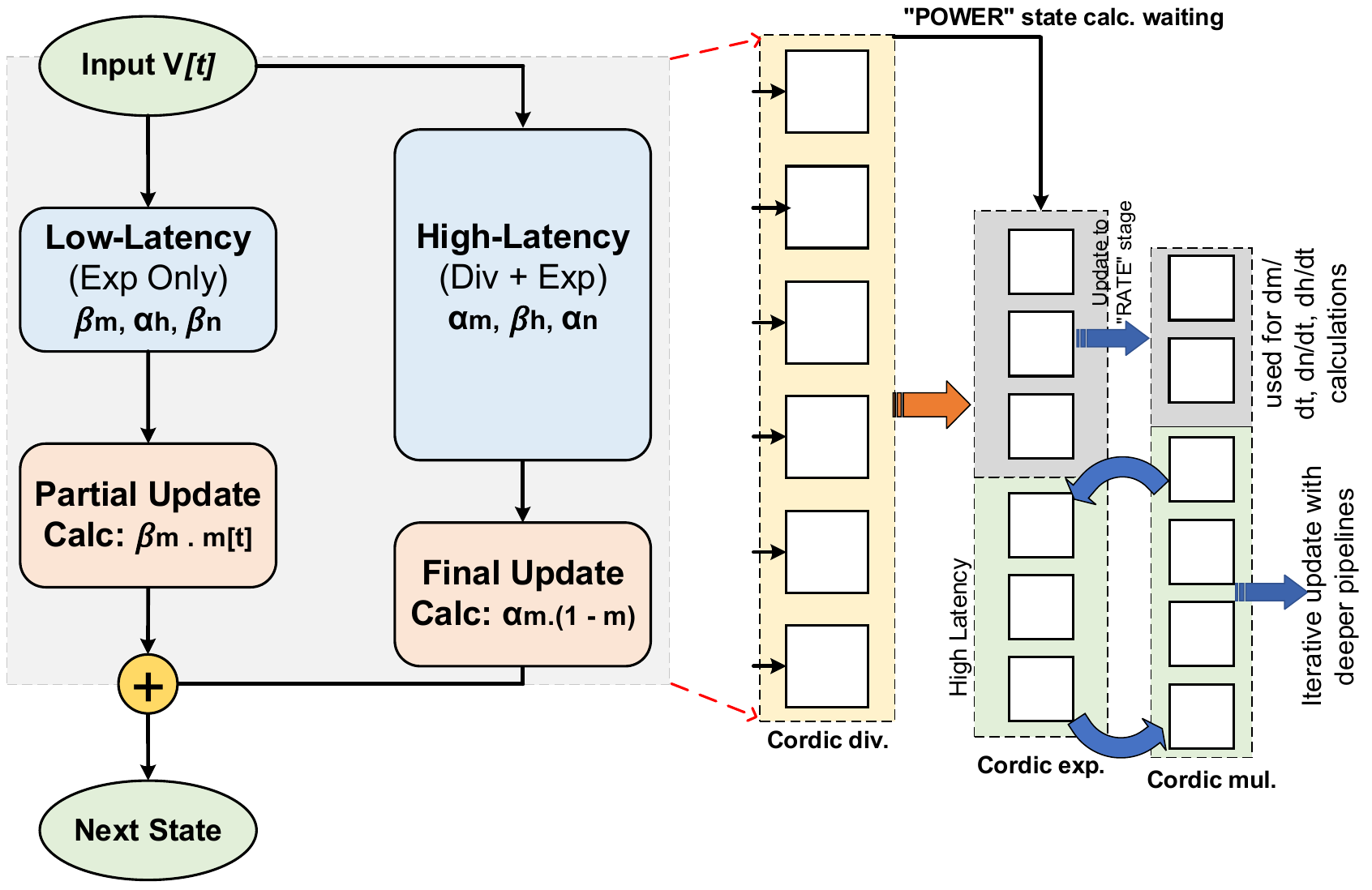}
    \caption{\textcolor{black}{Block diagram of CNP methodology with the CAMP Architecture and scheduling overview.}}
    \label{fig:block_diagram}
\end{figure}

\Copy{R1.4.1}{The approach ensures that the latency for division operations is hidden under the next processing stage, and results in architectural-level temporal complexity of order $O(1)$ relative to gate count by overlapping the Rate Calculation Phase with the State Update Phase. We achieved 34.7\% higher utilisation across all 18 cores with the Modular-Parallel approach, resulting in a 42.39\% speed-up. We ensure efficiency through a software-hardware co-design approach by tiling the data before feeding it to the CNP and finally stitching it back at the end, ensuring no CORDIC core is left stalling (waiting) at any stage; the pipeline depth (3-6 stages for High-Latency and Low-Latency groups, respectively) is fully saturated, ensuring maximum utilisation of the FPGA resources.}
}

\subsection{\textcolor{black}{Precision and Stability}}
\Copy{R1.2.4}{To address numerical instability where rate equation denominators approach zero ($-55$mV), ReLANCE integrates "Special Case" detectors. Upon detecting an $\epsilon$-neighbourhood ($\epsilon = 2^{-10}$) around singularities, the pipeline triggers L'Hôpital's rule approximations via shift-and-add logic. This prevents overflow and contributes to a 70\% improvement in NRMSE compared to standard fixed-point implementations.}

\begin{figure}[!t]
\centering
\subfloat[]{\includegraphics[width=0.75\columnwidth]{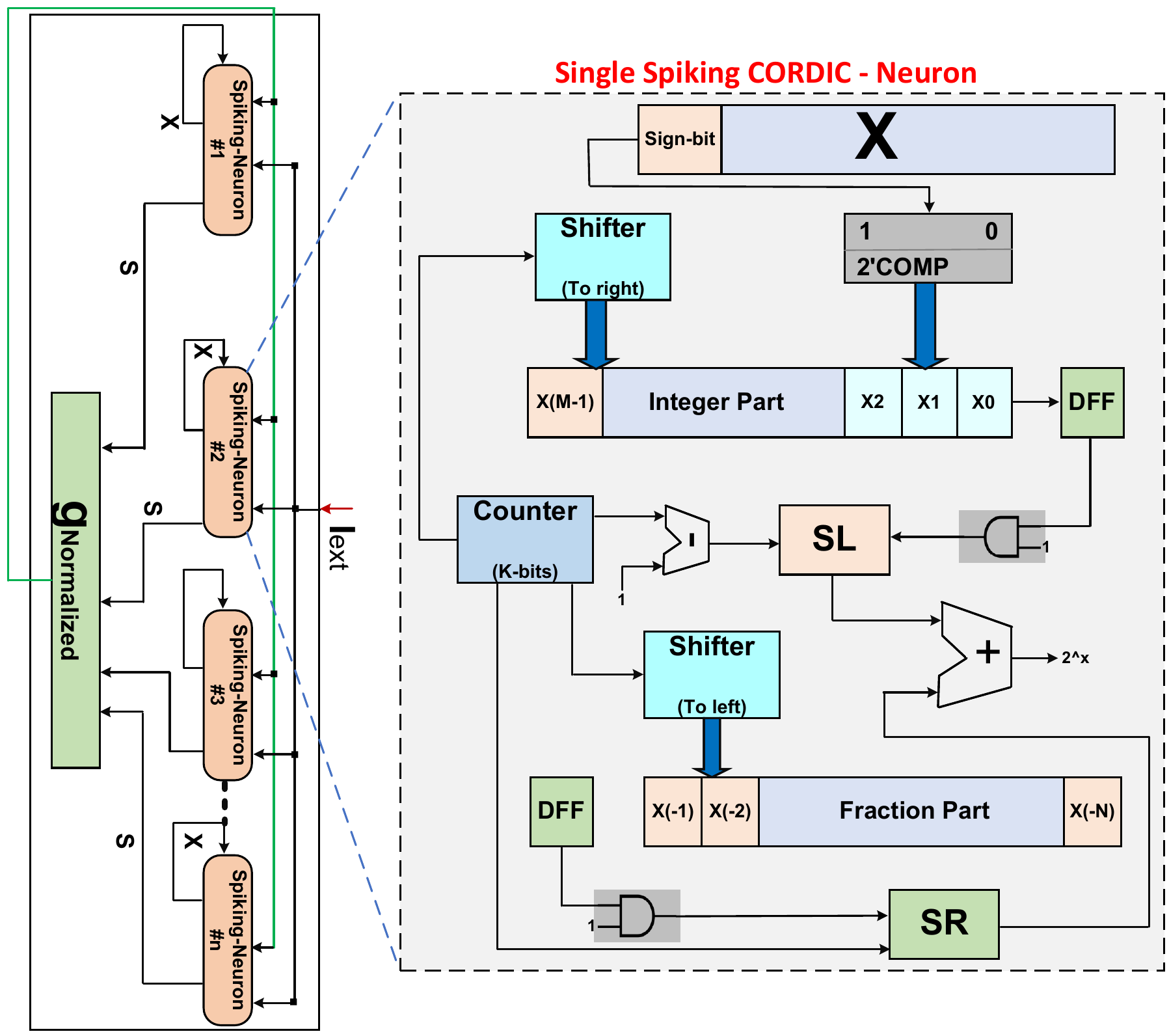}%
}
\hfill
\subfloat[]{\includegraphics[width=0.9\columnwidth]{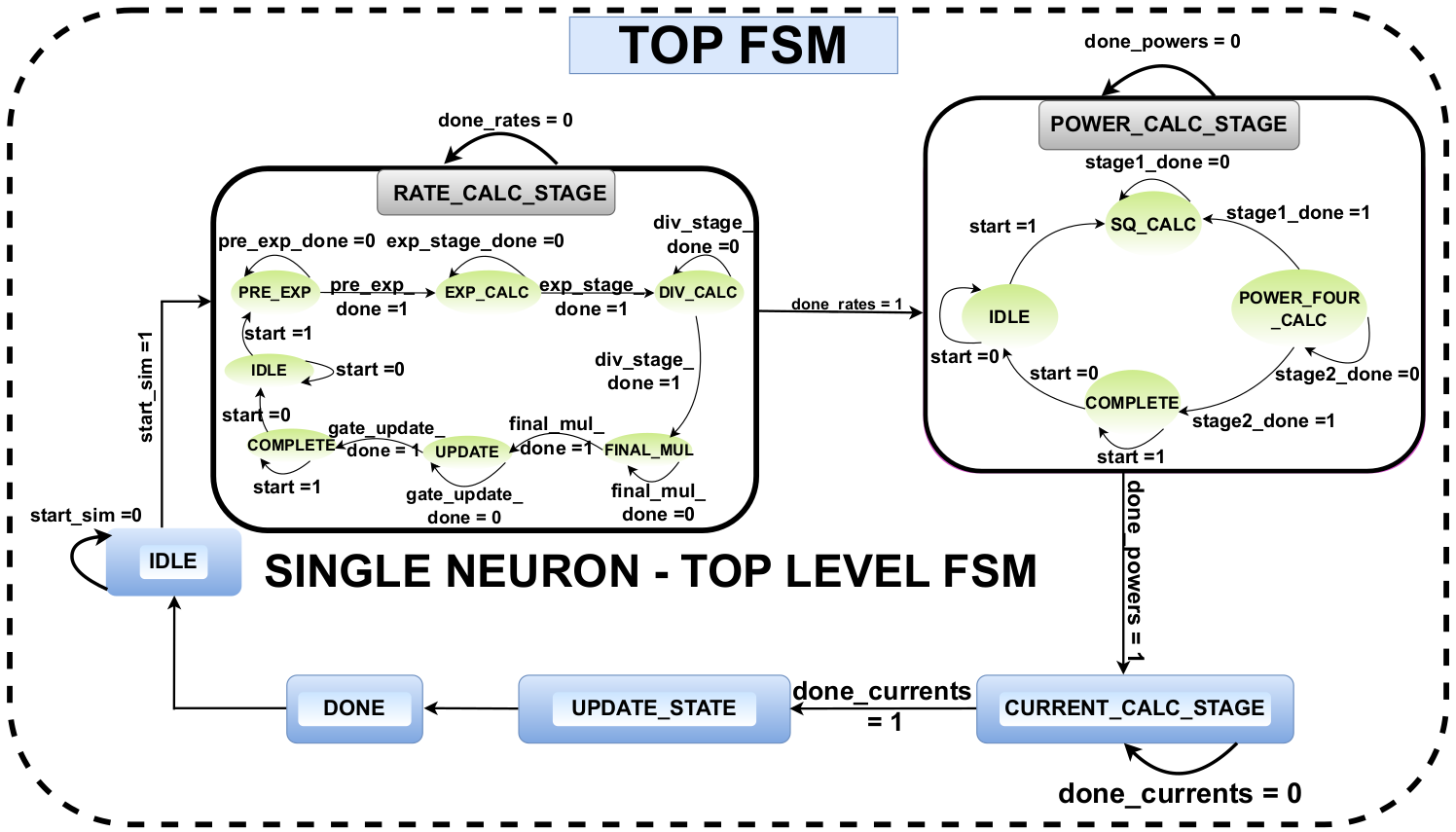}}%
\caption{(a) Architecture for the spiking neuronal pool, the functional equivalent of the DNN neuron engine for the execution of AI workloads with RCHH Neuron, and \textcolor{black}{(b) the complete top-level FSM of a single neuron, showing different neuron states.}}
\label{fig:RCHH-FSM}
\end{figure}

\begin{figure}[!t]
    \centering
    \includegraphics[width=0.875\columnwidth]{Fig/Spike-pattern.pdf}
    \caption{Diverse spiking characteristics generated by the RCHH model. Top: Fast Spiking: High-frequency non-adapting spikes (cortical neurons); Intrinsically Bursting: Spike clusters followed by hyperpolarisation (thalamic neurons); Chattering: Gamma-frequency oscillations (30–80 Hz) during sustained input. Bottom: Low-Threshold Spiking: Subthreshold oscillations leading to delayed firing; Tonic/Rebound Spiking: Persistent firing post-inhibitory input (thalamocortical cells).}
    \label{fig:spike-pattern}
\end{figure}

\begin{figure}[!t]
    \centering
    \includegraphics[width=0.85\columnwidth]{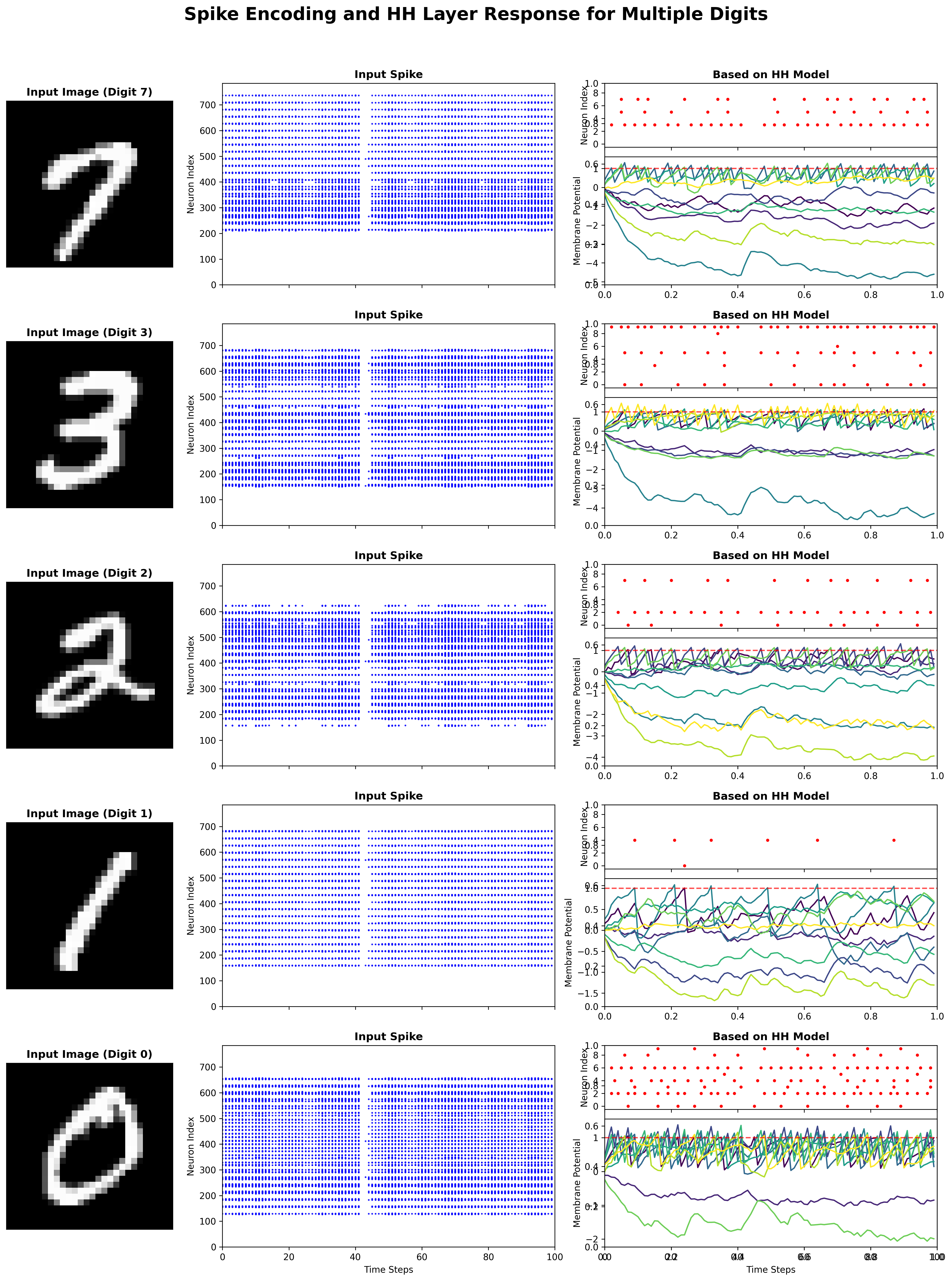}
    \caption{Network implementation with only 0.35\% accuracy drop: a $784$-neuron input layer receives rate-coded spikes from each $28\times28$ MNIST image. Synaptic weights learned via STDP generate the injected current $I_{\text{inj}}$, driving spiking activity in downstream neurons to produce the final digit prediction (the class with the maximum output spikes).}
    \label{fig:accuracy}
\end{figure}

Fig.~\ref{fig:spike-pattern} demonstrates our RCHH neuron model's ability to mimic distinct firing modes observed in biological neurons. These are critical for encoding spatiotemporal features in neuromorphic applications.

\section{Performance Analysis}
The experimental setup for evaluation consists of a quantised SNN evaluation based on the emulation of the CORDIC-based H\&H neuron, followed by a hardware implementation using Verilog HDL. We implemented the proposed Neuron and neuronal pool shown in Fig. \ref{fig:block_diagram}, and \ref{fig:RCHH-FSM} for evaluation using the VC707 FPGA. An FPGA board was chosen for a fair comparison with the state-of-the-art MAC unit, offering the same arithmetic precision. From the results in Table \ref{tab:neuron-fpga}, our proposed unit consumes significantly fewer LUTs and shows faster computation compared to other neuron types. \textcolor{black}{\Copy{R3.1}{The system with VC707 deployment inferred 5.04 ms latency and 26.34 mW, outperforming prior works with similiar parameters: FPGA VC707 deployments \cite{DNN_TRETS'23, Flex-PE} (21.46 ms / 2.24 W), Layer-reused (77.2 ms / 1.524 W), Pynq-Z2 \cite{CAL'17} (18.4 ms / 0.93 W ), and baselines: NVIDIA Jetson Nano (GPU only) (53.5 ms / 85.45 mW) and Raspberry Pi (CPU only) (75.5 ms / 102.8 mW).}} Although our RCHH neuron consumes more resources compared to the CORDIC-based iterative Izhikevich neuron, it has proven to be more precise and biologically accurate than the latter. We further extended the analysis for the MNIST pattern and evaluated against DNNs (LeNet-5/MNIST) for a fairer comparison. DNN performance for the CORDIC-based neuron model\cite{DNN_TRETS'23} was 95.06\% (8-bit), 96.70\% (16-bit) and 96.80\% (32-bit) respectively and 94.5\% in \cite{CAL'17}. The SNN-model performance with the proposed CORDIC-based H\&H neuron is reported in Fig.~\ref{fig:accuracy}. Further, the observation for spiking-transformer/ImageNet-1K is that accuracy drops from FxP-16 quantisation (0.75\%-1.2\%) and an additional 0.6\% from CORDIC RCHH configuration against the FP32 baseline.

\begin{table}[!t]
\centering
\caption{Comparison for FPGA Resources utilization,\\ with different 16-bit Neurons}
\label{tab:neuron-fpga}
\renewcommand{\arraystretch}{1.1}
\resizebox{0.75\columnwidth}{!}{%
\begin{tabular}{|l|c|c|c|}
\hline
\textbf{Design}& \textbf{LUT} & \textbf{FF} & \textbf{Delay (ns)} \\ \hline
\begin{tabular}[c]{@{}l@{}}Proposed\\ (Pareto CORDIC H\&H)\end{tabular} & 1770 & 862 & 1.4 \\ \hline
Iterative CORDIC H\&H\cite{HH-TCASI'21} & 2344 & 460 & 5 \\ \hline
CORDIC Izhikevich\cite{Izhikevich_NEWCAS'21} & 986 & 264 & 2.16 \\ \hline
Multiplier-less H\&H\cite{PWL_HH_Frontiers'14} & 5660 & 2840 & 11.77 \\ \hline
RAM H\&H\cite{PWL_HH_Frontiers'14} & 4735 & 1552 & 10 \\ \hline
RPE\cite{RPE} & 8054 & 1718 & 4.62\\\hline
MP-RPE\cite{RPE} & 8065 & 1072 & 5.56 \\\hline
PWL H\&H\cite{HH-TCASI'21} & 29130 & 25430 & 39.06 \\ \hline
Parallel CORDIC H\&H & 86032 & 50228 & 15.78 \\ \hline
\end{tabular}}
\end{table}

\textcolor{black}{Our design is relatively superior to state-of-the-art LIF and H\&H implementations in terms of NRMSE, NRMSD, and ERRT metrics: (0.13, 0.34, 0.32) compared to (0.43, 0.39, 0.21) and (0.93, 0.74, 5.00) in \cite{IF_TCASI'16, HH-TCASI'21}, achieving a CORR of 93.81\% against 95.66\% \& 90.4\%. This is also attributed to the special-case handling required for rate calculations, as well as to the near-zero denominator in division, which results in a lower error rate (ERRT).} For comparison with the DNN accelerator engine, we built an iso-functional neuronal pool with 64 neurons. Our design shows significantly reduced resource usage (LUTs and FFs) compared to prior SNN models and DNN accelerator engines, as shown in Table \ref{tab:neuron-fpga}.


\begin{table}[!t]
\centering
\caption{\Copy{R2.3.2}{Hardware resource comparison with state-of-the art accelerator designs, 16-bit precision, and VC707 FPGA.}}
\label{tab:neuron-fpga}
\renewcommand{\arraystretch}{1.1}
\resizebox{0.775\columnwidth}{!}{%
\begin{tabular}{|l|c|c|l|}
\hline
\textbf{Design} & \multicolumn{1}{c|}{\textbf{LUTs (K)}} & \multicolumn{1}{c|}{\textbf{FFs (K)}} & \multicolumn{1}{c|}{\textbf{Delay (ns)}} \\ \hline
\begin{tabular}[c]{@{}l@{}}Proposed\\ (Pareto CORDIC H\&H)\end{tabular} & 118.6 & 57.76 & 5.04 \\ \hline
Iterative CORDIC H\&H\cite{HH-TCASI'21} & 157 & 30.82 & 20.5 \\ \hline
CORDIC Izhikevich\cite{Izhikevich_NEWCAS'21} & 66 & 17.68 & 9.288 \\ \hline
Multiplier-less H\&H\cite{PWL_HH_Frontiers'14} & 359.2 & 190 & 31.54 \\ \hline
RAM H\&H\cite{PWL_HH_Frontiers'14} & 317.3 & 104 & 35.6 \\ \hline
TCAD'23\cite{TCAD'23} & 170.4 & 113.2 & 7.38 \\ \hline
TRETS'23 \cite{DNN_TRETS'23} & 115 & 115 & 21.46 \\ \hline
TCAS-I'22\cite{TCAS-I'22} & 213 & 352 & 5 \\ \hline
TCAS-II'23 & 132 & 39.5 & 6.68 \\ \hline
ESL'24\cite{ESL'24} & 195 & 95.7 & 5.36 \\ \hline
\end{tabular}}
\end{table}

\begin{figure*}
    \centering
    \vspace{-5mm}
    \subfloat[\label{fig:Throughput vs Hidden Width}]{\includegraphics[width=0.3\textwidth, height=42.5mm]{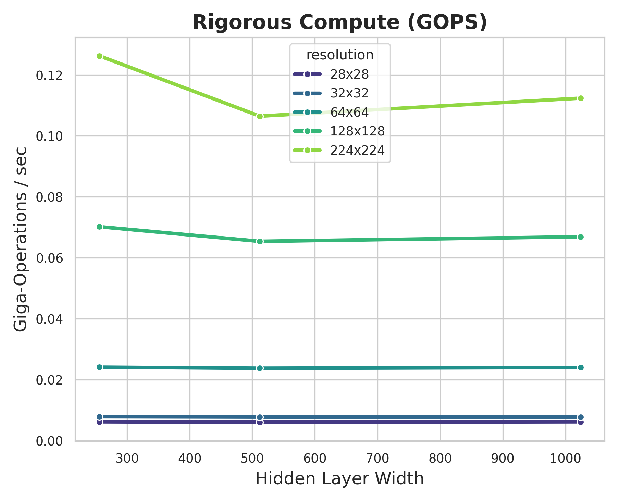}}
    \hfill
    \subfloat[\label{fig:MSOPs vs Input Width}]{\includegraphics[width=0.3\textwidth, height=42.5mm]{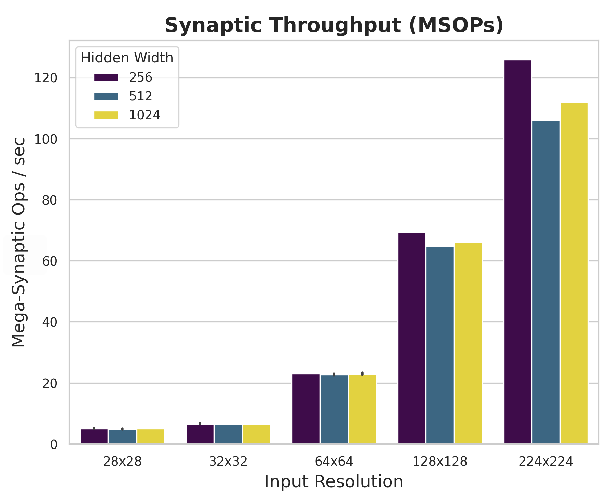}}
    \hfill
    \subfloat[\label{fig:Latency vs Hidden Width}]{\includegraphics[width=0.3\textwidth, height=42.5mm]{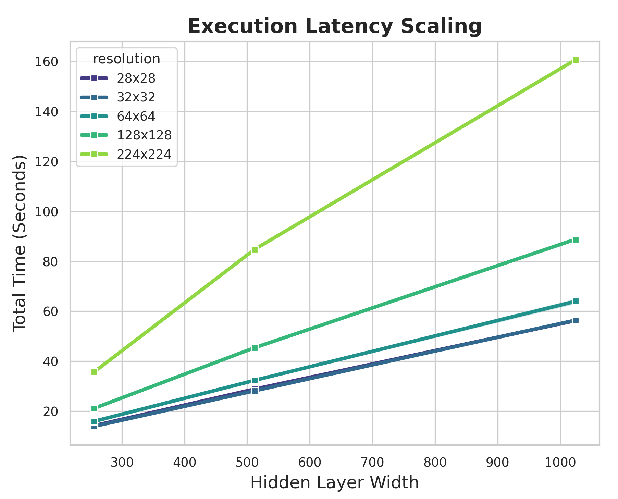}}
    \caption{(a) System Throughput (GOPS) vs. Hidden Width across varying Input Resolutions, (b) Synaptic Throughput (MSOPS): Mega Synaptic Operations per second vs Input Resolutions, (c) Execution Latency Scaling: Total Time (in seconds) vs. Hidden Layer Width across varying Input Resolutions}
\end{figure*}


\textcolor{black}{\Copy{R1.4.3}{The evaluation targets architectural scalability, as detailed with MNIST-scale end-to-end evaluation and further scaling. As shown in Fig. \ref{fig:MSOPs vs Input Width}, synaptic throughput required scales near-linearly from 10.3 Giga-Spiking ops. per second (GSOPS) for $28 \times 28$ to 236.8 GSOPS for $224 \times 224$, indicating efficient support for ImageNet-1K inputs without significant control or scheduling bottlenecks as the architecture is parameterized and follows scalable startgey. Fig. \ref{fig:Throughput vs Hidden Width} shows that compute throughput (GOPS) requirements, defined by counting fixed-point arithmetic primitives (add, multiply, divide, exponential) per neuron timestep, remain comparable as hidden layer width increases from 256 to 1024 neurons, demonstrating effective CAMP architecture for latency-critical rate computations. Correspondingly, Fig. \ref{fig:Latency vs Hidden Width} shows that despite a 5.2× increase in input dimensionality from 28$\times$28 to 64$\times$64, normalised latency improves by 4.8× (from $8.8 \mu s/s/pixel$ to $1.8 \mu s/s/pixel$) and further reaches $1.5 \mu s/s/pixel$. at $224 \times 224$, confirming optimization for high-resolution inputs}. \Copy{R3.5}{Thus, CNP Architecture reported a total energy evaluation of 26.324 mJ per inference run, showing significant improvements compared to existing works\cite{HH-TCASI'21}, \cite{Izhikevich_NEWCAS'21}, \cite{PWL_HH_Frontiers'14}.}}
\textcolor{black}{\Copy{R3.4}{Scalability beyond the evaluated neuron pool can be analysed analytically, each RCHH neuron maintains a constant number of state variables, leading to linear growth in register memory and AXI bandwidth with network size, while a fixed-depth, neuron-local FSM ensures constant per-neuron control complexity and avoids centralized bottlenecks, making scaling primarily memory-bandwidth-limited and amenable to tiled or multi-core neuromorphic systems.}}

\section{Conclusion}
In this work, we demonstrate a resource-efficient CORDIC-based H\&H neuron and evaluate it for the Cortical Neural Pool engine used for SNN acceleration. The proposed design significantly reduces resource consumption and latency while maintaining high biological accuracy, utilizing a performance-optimised CORDIC-based design. Thus, our design achieves 24.5\% lower LUT consumption, 35.2\% faster execution, and a 70\% improvement in NRMSE compared to prior work\cite{IF_TCASI'16}, emphasising improved energy efficiency. The CNP pool achieves 12.69 GOPS throughput, outperforming iso-functional DNN accelerator engines, with just 0.35\% accuracy degradation on MNIST classification. These overall results establish the feasibility of SNN accelerators as promising AI accelerators focused on low-power, resource-constrained edge AI systems. Future work involves evaluating application-level performance in autonomous driving.

\newpage

\bibliographystyle{myieeetr}
\bibliography{bib}


\end{document}